\documentclass[preprint,times,review,12pt,sort,compress,3p]{elsarticle}

\usepackage{amssymb}
\usepackage{amsmath}
\usepackage{hyperref}
\usepackage{longtable}
\usepackage{multirow}
\usepackage{array}
\usepackage{natbib}
\usepackage{xcolor} 

\usepackage{caption}
\usepackage{booktabs}
\journal{Pattern Recognition}

\begin{document}

\begin{frontmatter}



\title{DeepSelective: Interpretable Prognosis Prediction via Feature Selection and Compression in EHR Data}


\author[a,b]{Ruochi Zhang\fnref{equal}} 
\ead{zrc720@gmail.com}
\author[a,b]{Qian Yang\fnref{equal}} 
\ead{yangqian_job@163.com}
\author[a,b]{Xiaoyang Wang} 
\ead{xw388@drexel.edu}

\author[d]{Tian Wang} 
\ead{twangct@connect.ust.hk}
\author[a,b]{Qiong Zhou} 
\ead{2019200106@mails.cust.edu.cn}
\author[e]{Ziqi Deng} 
\ead{zdengax@connect.ust.hk}
\author[a,b]{Kewei Li} 
\ead{kwbb1997@gmail.com} 
\author[a,b]{Yueying Wang} 
\ead{yueyingwang0419@163.com}
\author[a,b]{Yusi Fan} 
\ead{fan_yusi@163.com}
\author[a,b]{Jiale Zhang} 
\ead{jialez22@mails.jlu.edu.cn} 
\author[a,b]{Lan Huang} 
\ead{huanglan@jlu.edu.cn} 
\author[c]{Chang Liu\corref{cor1}} 
\ead{liuchang@blsa.com.cn} 
\author[a,b]{Fengfeng Zhou\corref{cor1}} 
\ead{FengfengZhou@gmail.com} 

\cortext[cor1]{Corresponding authors.}
\fntext[equal]{These authors contributed equally to this work.}

\affiliation[a]{organization={College of Computer Science and Technology},
            addressline={Jilin University}, 
            city={Changchun},
            postcode={130012}, 
            country={China}}
\affiliation[b]{organization={Key Laboratory of Symbolic Computation and Knowledge Engineering of Ministry of Education},
            addressline={Jilin University}, 
            city={Changchun},
            postcode={130012}, 
            country={China}}
\affiliation[c]{organization={Beijing Life Science Academy},
            city={Beijing},
            postcode={102209}, 
            country={China}}
\affiliation[d]{organization={Department of Chemical and Biological Engineering},
            addressline={The Hong Kong University of Science and Technology}, 
            city={Clear Water Bay, Hong Kong},
            postcode={999077}, 
            country={China}}

\affiliation[e]{organization={School of Science},
            addressline={The Hong Kong University of Science and Technology}, 
            city={Hong Kong},
            postcode={999077}, 
            country={China}}
            
\begin{abstract}
The rapid accumulation of Electronic Health Records (EHRs) has transformed healthcare by providing valuable data that enhance clinical predictions and diagnoses. While conventional machine learning models have proven effective, they often lack robust representation learning and depend heavily on expert-crafted features. Although deep learning offers powerful solutions, it is often criticized for its lack of interpretability. To address these challenges, we propose DeepSelective, a novel end to end deep learning framework for predicting patient prognosis using EHR data, with a strong emphasis on enhancing model interpretability. DeepSelective combines data compression techniques with an innovative feature selection approach, integrating custom-designed modules that work together to improve both accuracy and interpretability. Our experiments demonstrate that DeepSelective not only enhances predictive accuracy but also significantly improves interpretability, making it a valuable tool for clinical decision-making. The source code is freely available at http://www.healthinformaticslab.org/supp/resources.php .
\end{abstract}




\end{frontmatter}



\section{Introduction}
\label{sec1}
In recent years, the rapid accumulation of Electronic Health Records (EHRs) and the widespread adoption of electronic medical information systems across various medical institutions have transformed the healthcare landscape \citep{nasarudin2024review}. EHRs capture essential information about laboratory tests and diagnostic decisions, fundamentally impacting the diagnostic process \citep{mulyani2024analysis}. This transformation brings substantial benefits but also introduces notable challenges that influence patient safety and the quality of healthcare \citep{sittig2025recommendations}. The vast repository of EHRs data can greatly enhance both quantitative prediction and subjective diagnosis in clinical practice \citep{4}. Quantitative prediction models provide valuable supplementary information, especially for less experienced clinicians, significantly helping to reduce misdiagnosis rates \citep{karaferis2025design}.

Both conventional machine learning and deep learning have been deployed in various clinical prediction tasks \citep{zhang2025brain}, but many computational challenges remain unresolved due to the complicated nature of human health status. Conventional machine learning models have been successfully used to answer some clinical diagnosis and prognosis questions based on the EHRs data but they provide limited capabilities in representation learning \citep{8}, and rely on the experts’ experience to hand-craft effective features like Body Mass Index (BMI). Deep learning fully utilizes the architecture of neural net-works and demonstrates a strong capability in representing the complex inter-feature patterns. However, lack of interpretability is one of the main obstacles to prevent deep learning from a broader deployment in the clinical practice \citep{11}. 

To address these challenges, we propose DeepSelective, a novel framework designed to predict patient prognosis using EHRs data while enhancing model interpretability. DeepSelective is built upon the principle that both generalization performance and interpretability are critical for practical healthcare applications. The contributions of this study can be summarized as follows: 
\begin{itemize}
\item  The study introduces DeepSelective, a comprehensive framework that significantly enhances both predictive accuracy and interpretability in prognosis prediction. This is achieved through the integration of innovative modules, including the Dynamic Gating Feature Selection (DGFS), the Attentive Transformer Autoencoder (ATA), and the Representation Matching Layer (RML), which synergistically optimize feature selection, data compression, and model generalization.

\item The effectiveness of the DeepSelective framework is thoroughly validated through rigorous experiments, demonstrating its superior performance in prognosis prediction. As a result, DeepSelective emerges as a valuable tool for clinical decision-making, offering reliable and interpretable insights into patient outcomes.
\end{itemize}

\section{Related Work}
\subsection{Feature Selection for EHRs}
\label{subsec1}
Feature selection is crucial for building predictive models from EHRs, traditionally relying on labor-intensive expert-defined features \citep{12}. Deep learning has shifted this paradigm by enabling automatic feature construction \citealp{zhang2024multi}, but often lacks interpretability \citep{16}. On the contrary, feature selection step ensures the interpretability of a machine learning model by explicitly recommending biologically meaningful biomarkers for training the prediction models \citep{13}. Recent approaches, such as Gronsbell et al.’s automated feature selection method \citep{12} and Liu et al.’s EnRank algorithm \citep{19}, aim to balance accuracy with interpretability by integrating biologically meaningful biomarkers.

\subsection{Predictive Modeling with EHRs}
Traditional machine learning techniques like logistic regression and decision trees have been widely used in healthcare for predictive modeling \citep{20}. These models, though effective, often require extensive manual feature engineering due to the complexity of EHRs data \citep{21}. \citep{23} proposed the e-Health system to predict fetal health, consisting of the tree-based feature selection algorithms and 9 binary classification models trained by the selected features. \citep{24} created the individual models to predict the possibility of being re-admitted in the future using the features recommended by the Correlation-based Feature Selection (CFS) method. Many more studies may be found in the literature, including the recent review \citep{25}. Deep learning offers robust capabilities for learning complex inter-feature patterns, eliminating the need for manual feature selection \citep{26}. Models like AdaCare and GRASP enhance healthcare predictions by learning from large datasets or similar patient data, respectively \citep{26}. However, the black-box nature of deep learning presents challenges in interpretability, which is crucial for clinical adoption \citep{28}.

\subsection{Interpretable Deep Learning Approaches}
To enhance the interpretability of deep learning models, some studies integrate traditional interpretable modules \citep{li2022interpretable}. For instance, the wide \& deep learning framework combines linear models with deep neural networks to improve both prediction accuracy and interpretability \citep{32}. Other methods, such as NeuralFS and Cancelout, focus on feature selection within neural networks, though they don’t fully leverage deep learning’s representation learning capabilities \citep{33}. More recently, ISP-IRLNet proposes a joint optimization framework that combines an interpretable sampling module with an implicit regularization network, showing that interpretability can also be embedded in data acquisition and reconstruction processes, further broadening the scope of interpretable deep learning\citep{li2024isp}.

\section{Background}

\subsection{Attention Mechanism}
Attention mechanisms have become a cornerstone in sequential data modeling, particularly for tasks that involve processing time-series data, natural language, or any form of sequential input where the context and the relationships between elements are crucial. Attention gives different weights to the input sequences, allowing the networks to focus on the relevant part as needed. Following expression in matrix form introduced by \citep{vaswani2017attention}, the attentive value is computed by
\begin{equation}
\text { Attention }(Q, K, V)=\operatorname{softmax}\left(\frac{Q K^{T}}{\sqrt{d_{k}}}\right) V
\end{equation}
where $Q$, $K$ and $V$ are named query, key and value matrix which are learnable parameters. $d_k$ stands for the dimension of key vectors and the normalization by $\sqrt{d_{k}}$ leads to a stable gradient in the training process. The attention mechanism is furthermore constructed by performing the above attention function several times, namely multi-head attention. 

\subsection{Gumbel-Softmax}\label{sec: gumbel-softmax}
The Gumbel-Softmax trick is a powerful tool in deep learning that bridges the gap between discrete and continuous optimization landscapes \citep{jang2016categorical}. Providing a differentiable approximation to sampling from categorical distributions, it opens up new possibilities for training deep learning models that involve discrete variables.
Let $z = [z_1, z_2, \dots, z_d] $ be a one-hot stochastic variable and $\pi = [\pi_1,\pi_2, \dots \pi_d]$ be the corresponding class probabilities. The Gumbel-Softmax trick provides a continuous approximation to the categorical distribution, denoted by 
\begin{equation} \label{gsoftmax}
z_i = \frac{\exp{((\log\pi_i + g_i)/{\tau})}}{\sum_{j = 1}^{d} \exp{((\log\pi_j + g_j)/{\tau})}}
\end{equation}
where $g_1$, $g_2$,$\dots$,$g_d$ are i.i.d. samples drawn from $\text{Gumbel}(0, 1)$. $\tau$ is the temperature parameter controlling the smoothness of the approximation. As $\tau \rightarrow 0$, the softmax function becomes a hardmax, and the approximation $z_i$ converges to either 0 or 1, replicating the behaviour of a true categorical variable, i.e.

\begin{equation}
\lim_{\tau \rightarrow 0}z_i =\left\{\begin{array}{l}
1 \ \ \ \text { if } i =\operatorname{\arg\max}_j(\log \pi_j + g_j) \\
0 \ \ \ \text { otherwise }
\end{array}\right.
\end{equation}

One of the most important advantages of leveraging Gumbel-Softmax is enabling gradient-based optimization for models with discrete outputs. Compared to the "$\arg\max$" operation, it offers a more stable training process by avoiding discontinuity in the loss function.

\section{Methodology}
\subsection{Problem Formulation}

The content of EHRs changes as the patient’s health status evolves. It is usually organized as a series of time-ordered records, denoted as $\mathbf{X} \in \mathbb{R}^{T \times N}$. Note that $\mathbf{X}$ is a time-dependent matrix and it is indeed the concatenation of N-dimensional medical features $\mathbf{X}$ until the time of $t$, $i.e.$ $\mathbf{X}_t = [\mathbf{x}_1, \mathbf{x}_2, \dots, \mathbf{x}_t]$. Each electronic health record consists of medical information such as age, gender symptoms, etc. 
The prediction task in this study is formulated as the following description. Given the historical EHRs data entry of the patient, we need to predict this patient’s future health status $\mathbf{y_t}$ (also known as prognosis). This is defined as the probability of suffering from a specific risk in the future (e.g., decompensation). Given the patient's EHRs data at current time point, the modeling of EHRs data aims at predicting a patient’s health status in the future:
\begin{equation}
\mathbf{y_t} = \mathcal{F}(\mathbf{X_{t-1}})
\end{equation}
where $\mathcal{F}$ is an implicit function which can be approximated by deep neural networks. 

\subsection{The Proposed DeepSelective Framework}

DeepSelective is created as a versatile framework to achieve the above-mentioned prediction and enhance the interpretability of $\mathcal{F}$. We enhance interpretability through feature selection, enabling the model to explicitly indicate which features the output $\mathbf{y_t}$ depends on. In developing a predictive model for EHRs data, the foundational intuition stems from the principle that both the enhancement of generalization performance and interpretability hinge critically on effective data compression. This understanding guide the creation of the dual-module network comprising the Dynamic Gating Feature Selection module (DGFS) and the Attentive Transformer Autoencoder (ATA). Let's start by defining two concepts.

\begin{itemize}
    \item  \textbf{Sparsity Compression}: a lossy compression method, it enhances model interpretability by selecting only the most significant features, reducing the dimensionality.
    \item \textbf{Perceptual Compression:} a lossless compression method, it improves model prediction performance, especially generalization capability, by perceiving and compressing the most critical features using attention mechanisms and an encoder-decoder architecture.
\end{itemize}

To achieve generalization and better interpretability under the condition of a limited dataset, we need to find a balance between sparsity compression and perceptual compression.
The DGFS module is pivotal for sparsity compression, which is achieved by selecting the most salient $k$ features. This approach not only decreases data complexity but also preserves the most influential characteristics, thereby improving the model's performance and interpretability within a specified reduced space.

Conversely, the ATA module focuses on perceptual compression through a tailored attention mechanism. This approach allows the model to dynamically identify and focus on the most critical features from the compressed input provided by DGFS. By integrating these features into an encoder-decoder structure, ATA constructs a refined latent space $D^r$,which is essential for improving prediction accuracy while maintaining the model's ability to provide meaningful insights into the data's underlying patterns.

To leverage the compressed representations of DGFS and ATA module, we introduce additionally the RML module aiming to maximize the consistency between the distributions of the two low-dimensional structures and to extract complementary information from both representations.

\begin{figure}
    \centering
    \includegraphics[width=1\linewidth]{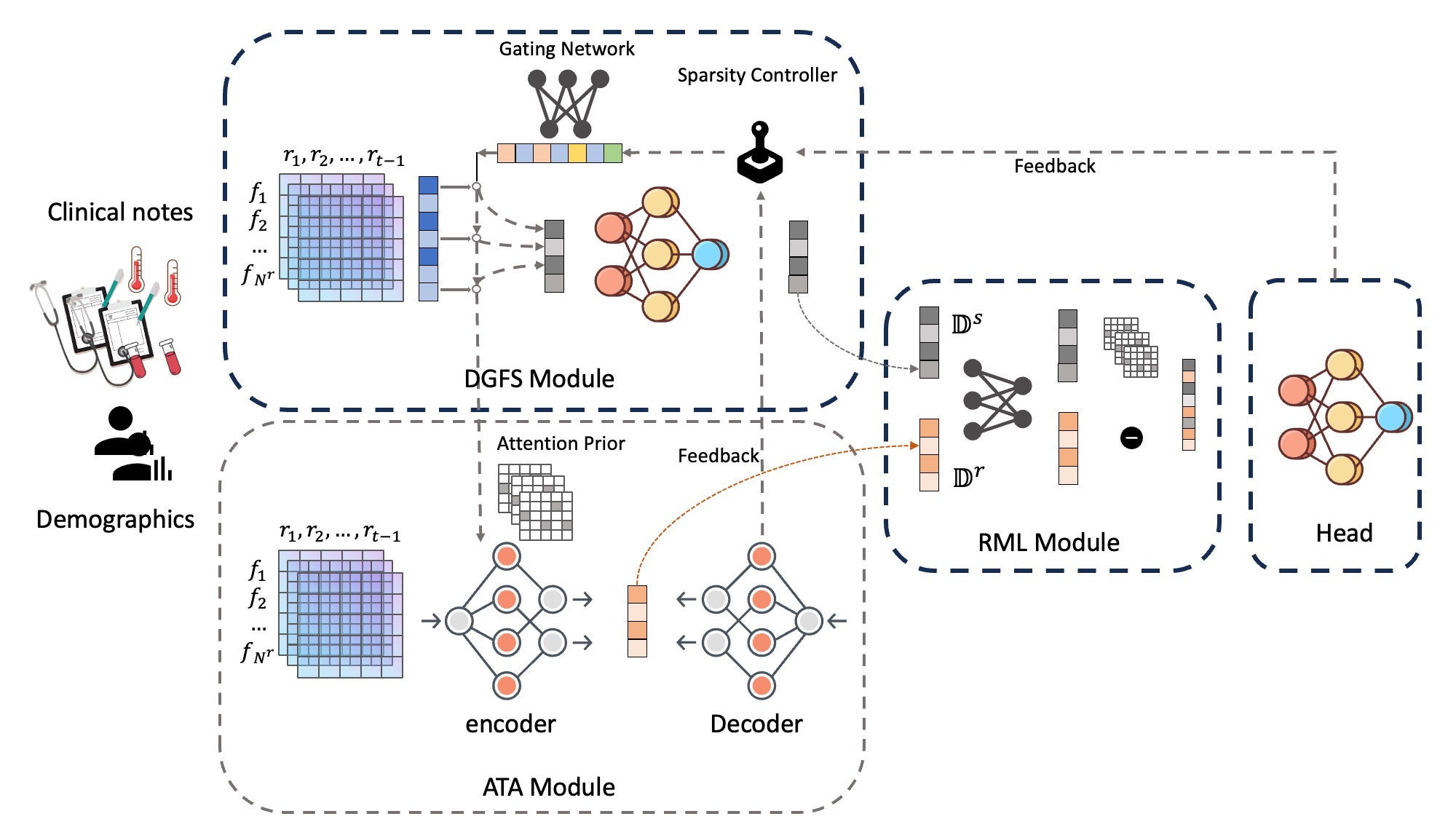}
    \caption{The overall architecture of DeepSelective.}
    \label{fig:f1}
\end{figure}

Therefore, the overall predictive problem can be formulates as,

\begin{equation}
\mathbf{\hat{y_t}}, \mathcal{S} = \hat{\mathcal{F}}(\mathbf{X_{t-1}})
\end{equation}

where $\hat{\mathcal{F}}$ stands for the DeepSelective network, $\mathbf{\hat{y_t}}$ is the predicted prognosis at time $t$ and $\mathcal{S}$ is a subset of the feature indices, addressing the key features in EHRs data.
\hyperref[fig:f1]{Fig.~\ref{fig:f1}} illustrates how the three modules, DGFS, ATA and RML, are orchestrated in the proposed DeepSelective framework.

\subsection{The Architecture of DGFS Module}
\label{sec: DGFS}

The Dynamic Gating Feature Selection (DGFS) module aims to explicitly select the most informative subset of features from high-dimensional EHR data while maintaining a balance between sparsity and predictive accuracy. DGFS applies a Gumbel-Softmax trick to approximate discrete selection during backpropagation, thus enabling end-to-end training. Below, we describe the DGFS mechanism in detail and explain why it does not cause data leakage into the subsequent \textit{Attentive Transformer Autoencoder (ATA)}.

To select the most informative subset of features, we want to minimize the prediction error while penalizing the number of utilized features. Mathematically, this is equivalent to minimizing the loss function below. 
\begin{equation}
L_{pred}(\mathbf{\theta}, \mathbf{w})=\|\mathbf{y} - {f}^{\theta}(\mathbf{X} \cdot \text{diag}(\mathbf{w}))\|_2 +\alpha \|\mathbf{w}\|_{0}
\end{equation}
where $f^{\theta}$ is a parameterized neural network, which will be elaborated upon in the following section. $\mathbf{w} \in \{0,1\}^{N}$ is a binary vector that controls how active are the features. The $L_0$ norm on $\mathbf{w}$ minimizes the number of active features in the model, enforcing the sparsity. However, $L_0$ optimization is challenging to directly apply to neural networks due to its complexity and non-convexity. 

\textbf{Sparse Selection.} We design the DGFS module leveraging Gumbel-Softmax approximation to bypass this issue. It tackles the problem from the probabilistic perspective and provides a soft selection of variables.
In our approach, the variable $\mathbf{p}$, computed via Gumbel-Softmax in section \ref{sec: gumbel-softmax}, represents the selection probability distribution of features. 

During training, to ensure that multiple features are selected, a hard thresholding is made on $\mathbf{p}$. We calculate the cumulative distribution function and select the smallest set of indexes whose cumulative probability exceeds 0.5. Mathematically, the set of activated indexes is defined as 

\begin{equation}
\begin{array}{cl}
\mathcal{S} = & \underset{}{\operatorname{\arg\min}} \operatorname{card}(\mathcal{A}) \\
& \text {s.t.} \sum_{i \in \mathcal{A}}\mathbf{p}_i \geq 0.5
\end{array}    
\end{equation}

The choice of the threshold value 0.5 is based on the principle of ensuring that a significant portion of the probability mass is captured while maintaining sparsity. The $|\mathcal{S}|$-dimensional sub-matrix with column indexes in set $\mathcal{S}$ is written as
\begin{equation}
\mathbf{X}_{\mathcal{S}}=[\mathbf{x}_i], i \in \mathcal{S}    
\end{equation}


We avoid treating sparsity strength $\tau$ as a fixed hyperparameter, as it should be adjusted based on data characteristics. To achieve this, we introduce the sparsity control module, which dynamically adjusts $\tau$ using a feedback-based approach inspired by the PID (Proportional-Integral-Derivative) controller \citep{willis1999proportional}. The sparsity controller considers past errors, reacts to the current error and predicts future error trends. This approach ensures that $\tau$ is adjusted according to both the prediction error and the reconstruction error from the ATA model, helping to balance between feature selection and information retention. The dynamic adjustment of $\tau$ can be formulated as:

\begin{equation}
\tau_{t+1} = \tau_t + k_p e_t + k_i\sum_{i=0}^t e_i + k_d(e_t - e_{t-1})
\end{equation}

where $e_t$ is the sum of prediction error and alignment error at time $t$ with coefficients $k_p$, $k_i$, $k_d$ controlling the proportional, integral and derivative terms respectively. 
The introduction of sparsity by the DGFS module is crucial as it forms the cornerstone of the interpretability of our entire network. 


\subsubsection{Avoiding Information Leakage}
A crucial design goal is to avoid “information leakage,” $i.e.$, preventing ATA from inadvertently accessing unselected features. In our framework, ATA only operates on \emph{masked features} $\mathbf{x}_S = \mathbf{x} \odot \mathbf{m}$, where $\odot$ denotes element-wise product. Thus, all unselected features are effectively zeroed out. 

Formally, let $\mathcal{L}$ be the model’s overall training loss. Suppose $\mathbf{x}_{\neg S}$ are the unselected features. The key to \emph{no leakage} is showing that:
\begin{equation}
    \frac{\partial \mathcal{L}}{\partial \mathbf{x}_{\neg S}} \;=\; \mathbf{0}.
    \label{eq:no-leak}
\end{equation}
Because ATA (and subsequent layers) only receive $\mathbf{x}_S = \mathbf{x} \odot \mathbf{m}$, any gradient path from $\mathbf{x}_{\neg S}$ to ATA’s output is cut off by the multiplication with the mask. Concretely, for an arbitrary output $h$ of ATA, we have:
\begin{equation}
    h \;=\; \mathrm{ATA}\bigl(\mathbf{x}_S\bigr) 
        \;=\; \mathrm{ATA}\bigl(\mathbf{x} \odot \mathbf{m}\bigr).
\end{equation}
Taking partial derivatives,
\begin{equation}
    \frac{\partial h}{\partial x_i} \;=\; 
    \frac{\partial \mathrm{ATA}(\mathbf{x} \odot \mathbf{m})}{\partial (\mathbf{x} \odot \mathbf{m})}
    \;\times\;
    \frac{\partial (x_i \cdot m_i)}{\partial x_i}.
\end{equation}
For any unselected feature $i \in \neg S$, $m_i = 0$. Hence,
\begin{equation}
    \frac{\partial (x_i \cdot m_i)}{\partial x_i} \;=\; m_i \;=\; 0,
\end{equation}
which implies 
\begin{equation}
    \frac{\partial h}{\partial x_i} = 0, 
    \;\;\;\; \forall \, i \in \neg S.
\end{equation}
Because ATA’s downstream losses ($e.g.$, reconstruction or alignment) solely depend on $h$, any partial derivative of the total loss $\mathcal{L}$ with respect to $x_i$ for $i \in \neg S$ is also zero, proving \hyperref[eq:no-leak]{Eq.~\eqref{eq:no-leak}}. Thus, there is no gradient flow from the unselected features into the loss, guaranteeing that \emph{no information is leaked} to the ATA module.

\subsection{The Architecture of ATA Module}

ATA utilizes a transformer-based autoencoder to achieve end-to-end data compression which specifically focus on the features selected by DGFS. 
The encoder is a stacked structure of multiple transformer layer which is enhanced based on attention mechanisms adjusted by the importance scores from DGFS:
\begin{equation}
    \text{Attention}(Q_{\mathcal{S}},K_{\mathcal{S}}, V_{\mathcal{S}}) = \text{softmax}\left(\frac{Q_{\mathcal{S}}K_{\mathcal{S}}^T}{\sqrt{d_k}}\right )V_{\mathcal{S}}
\end{equation}

$\mathcal{S}$ represents the set of selected supports in the DGFS module, deriving from the Gumbel-Softmax. 
The self-attention is performed on the selected support of original $K$, $Q$ and $V$ vectors, allowing the neural network to discard unimportant feature information.
Employing Gumbel-Softmax attention mechanism effectively elevates the influence of critical features within the latent representation, which not only boosts predictive accuracy but also enhances model interpretability. Consequently, the attention weight is allotted to the vital features, ensuring their significance is aptly reflected in the model's output. This is crucial since we want the networks to be highly dependent on interpretable features.

The decoder is also a multi-layer stacked transformer, but it does not adjust attention weights based on DGFS's output. This is because we want the decoder to reconstruct the input based on information in the latent space. In this module, we utilize Mean Squared Error (MSE) to calculate the pair-wise reconstruction error between the input and output to achieve lossless compression
\begin{equation}
\mathcal{L}_{ATA}(\mathbf{X}, \theta)=\| \mathbf{X} - \hat{\mathbf{X}} \|_{F}    
\end{equation}

\subsection{The Architecture of RML Module}

The Representation Matching Layer (RML) is a crucial component of DeepSelective, designed to synchronize the feature representations learned from DGFS and ATA while extracting complementary information. The core challenge in combining these representations lies in ensuring that the discrete feature selection (DGFS) aligns well with the continuous feature compression (ATA) without introducing redundancy or inconsistency. To address this, RML employs two key operations: $\mathbf{r}_{\text{add}}$ and $\mathbf{r}_{\text{sub}}$, which capture aligned and complementary information, respectively.

\subsubsection{Aligned Information: $\mathbf{r}_{\text{add}}$}
The aligned representation $\mathbf{r}_{\text{add}}$ is formulated as:
\begin{equation}
    \mathbf{r}_{add} = \sigma(\mathbf{W}_{1}\mathbf{z} + \mathbf{b}_{1})\odot(\mathbf{z}_s+\mathbf{z}_r)
\end{equation}

Here, $\mathbf{z}$ denotes the concatenation of the compressed representation from DGFS and the compressed representation from ATA. $\mathbf{W}$ and $\mathbf{b}$ are trainable parameters of a linear transformation that scales and shifts the concatenated features to facilitate effective integration. The sigmoid function acts as a gating mechanism that controls the contribution of each feature in the combined representation. The element-wise product ensures the contributions of $\mathbf{z}_s$ and $\mathbf{z}_r$ to the final representation $\mathbf{r}_{add}$ are weighted by their relevance as determined by the sigmoid gate. This formulation ensures that the combined representation is modulated by a learned relevance score, which allows the model to adaptively control the degree to which each feature contributes to the aligned representation. 

\subsubsection{Complementary Information: $\mathbf{r}_{\text{sub}}$}
Complementary information refers to unique or additional insights that one representation provides over the other. Extracting and integrating this information can lead to a more robust and comprehensive feature set for prediction. The complementary information $\mathbf{r}_{\text{sub}}$ is defined as:

\begin{equation}
\mathbf{r}_{sub} = \text{ReLU}(\mathbf{W}_2(\mathbf{z}_r-\mathbf{z}_s) + \mathbf{b}_2)
\end{equation}

The difference $\mathbf{z}_r-\mathbf{z}_s$ captures unique aspects of the ATA and DGFS representations, highlighting information that is present in one but not the other. Applying a ReLU activation ensures that only positive complementary contributions are retained, preventing cancellation effects.

The prediction of label $\mathbf{y}$ relies on combination of $\mathbf{r}_{sub}$ and $\mathbf{r}_{add}$.
Precisely, the parameterised function $f^{\theta}$ in the loss function in section \ref{sec: DGFS} is a multilayer perception that takes concatenation version of $\mathbf{r}_{sub}$ and $\mathbf{r}_{add}$ and outputs the estimated label of $\mathbf{y}$.
This concatenation form ensures that the final representation leverages both the aligned and complementary aspects of the feature sets from DGFS and ATA.

In addition, to ensure the consistency of the two compressed representations, $\mathbf{z}_s$ and $\mathbf{z}_r$, while preserving their complementary nature, we use cosine similarity to measure their similarity. The advantage of this approach is that it only considers the directional consistency of the representations, allowing for differences in their specific values. This is important because DGFS may emphasize selecting the most important features, while ATA aims to encode as much information as possible into a compressed representation. Thus, cosine similarity is an appropriate metric to achieve this balance. Therefore, we design the following loss function to minimize the directional difference between $\mathbf{z}_s$ and $\mathbf{z}_r$.

\begin{equation}
    \mathcal{L}_{align}(\mathbf{z}_r, \mathbf{z}_s) = \frac{\sum_{i=1}^n \mathbf{z}_{r,i} \mathbf{z}_{s,i}}{\sqrt{\sum_{i=1}^n \mathbf{z}_{r,i}^2} \sqrt{\sum_{i=1}^n \mathbf{z}_{s,i}^2}}
\end{equation}

Minimizing \( \mathcal{L}_{align} \) forces $\mathbf{z}_s$ and $\mathbf{z}_r$ to align in the latent space, ensuring consistency in learned representations.

\subsection{Final Representation and Loss Function}

The final representation \( r_{\text{final}} \) is obtained by concatenating \( r_{\text{add}} \) and \( r_{\text{sub}} \):
\begin{equation}
r_{\text{final}} = [r_{\text{add}}; r_{\text{sub}}]
\end{equation}

The model prediction is computed as:
\begin{equation}
\hat{y} = f_{\theta}(r_{\text{final}})
\end{equation}
where \( f_{\theta} \) is a multilayer perceptron (MLP) with trainable parameters \( \theta \).

The overall loss function combines prediction loss, alignment loss, and reconstruction loss:
\begin{equation}
L = L_{\text{pred}} + \beta_1 L_{\text{align}} + \beta_2 L_{\text{ATA}}
\end{equation}
where:
\begin{itemize}
    \item \( L_{\text{pred}} \) is the classification loss (e.g., cross-entropy loss).
    \item \( L_{\text{align}} \) ensures consistency between DGFS and ATA representations.
    \item \( L_{\text{ATA}} \) is the reconstruction loss from the autoencoder.
    \item \( \beta_1, \beta_2 \) are tunable hyperparameters.
\end{itemize}

\section{Experiments}


\subsection{Baselines}
This study compares the proposed DeepSelective framework with several baseline models that have been effective for EHRs data. These include GRU$_{\alpha}$ (attention-enhanced Gated Recurrent Units), RETAIN (a two-level neural attention model) \citep{35}, T-LSTM (an LSTM variant for handling irregular time intervals) \citep{36}, SAnD$^{*}$ (a self-attention-based model for clinical time-series data) \citep{38}, AdaCare (a health status model using dilated convolution and GRU) \citep{26} and ConCare (a model using self-attention for time-aware distribution of feature sequences) \citep{39}.

\subsection{Data Description and Evaluation Metrics}
Our research utilizes the MIMIC-III dataset, which contains EHRs data from over 90 ICUs at Beth Israel Deaconess Medical Center in Boston, Massachusetts \citep{42}. We follow the data cleaning, preprocessing, and partitioning methods outlined by \citep{44}. This process produces two benchmark datasets: one for early-stage ICU admission mortality prediction and another for real-time decompensation detection. 

We evaluate our model by the area under the precision-recall curve (AUPRC), the area under the receiver operating characteristic curve (AUROC), F1-Score and the minimum of sensitivity and precision Min(Se, P+) \citep{10}. To further evaluate the relevance and informativeness of the selected features, we utilize Mutual Information (MI) \citep{46}. MI measures the amount of information obtained about one random variable through another random variable. It helps quantify the dependency between the input features and the compressed features. The formula for mutual information between two random variables $X$ and $Y$ is given by:

\begin{equation}
I(X,Y) = \sum_{x \in X}\sum_{y \in Y}p(x,y)\log\left( \frac{p(x,y)}{p(x)p(y)} \right)    
\end{equation}

where $p(x,y)$ is the joint probability distribution of $X$ and $Y$, and $p(x)$ and $p(y)$ are the marginal probability distributions of $X$ and $Y$ respectively.

We also conduct T-test to statistically evaluate the significance of the selected features \citep{47}. The T-test assesses whether the means of two groups are statistically different from each other. In our context, the null hypothesis $H_0$ is that the mean importance of the selected features is equal to the mean importance of the non-selected features, while the alternative hypothesis $H_1$ is that the mean importance of the selected features is different from the mean importance of the non-selected features.

\subsection{Effectiveness of Compression}
\begin{figure}
    \centering
    \includegraphics[width=1\linewidth]{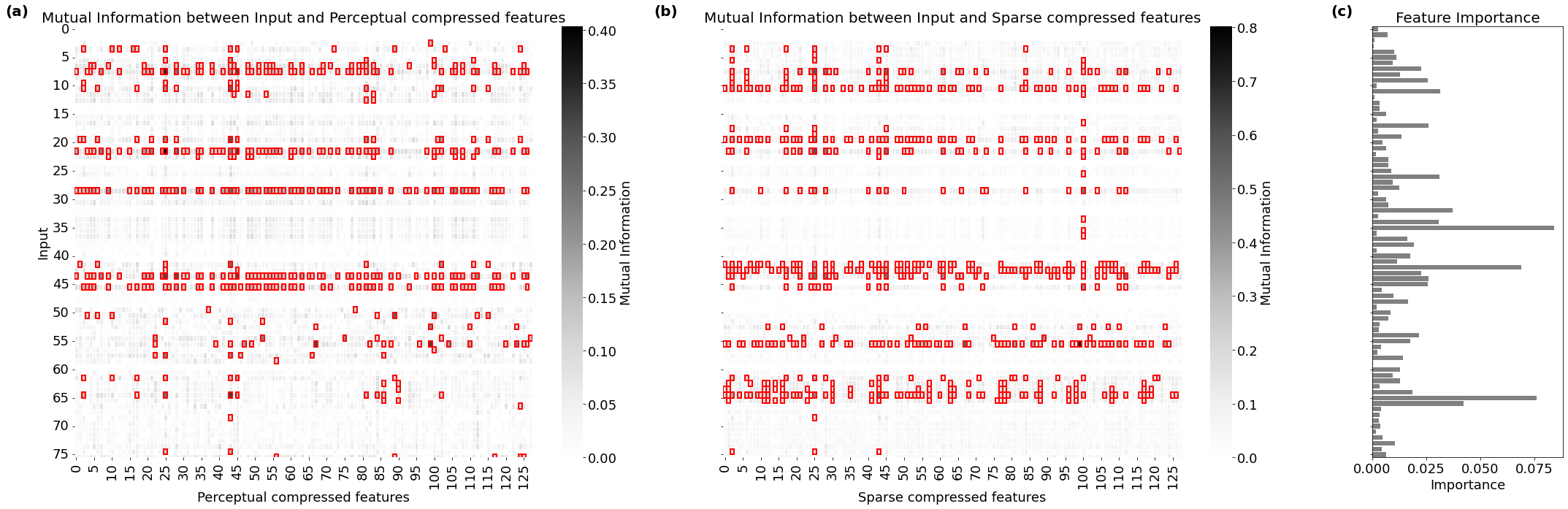}
    \caption{Mutual information analysis and feature importance evaluation for the proposed framework on mortality prediction task}
    \label{fig:f2}
\end{figure}

In this section, we evaluate the compression and feature selection
performance of the proposed framework, focusing
on the mutual information between input features and compressed
features. The results of these experiments are illustrated in \hyperref[fig:f2]{Fig.~\ref{fig:f2}}. \hyperref[fig:f2]{Fig.~\ref{fig:f2}(a)} shows the mutual information between input features and perceptual compressed features on mortality prediction task. (b) displays the mutual information between input features and sparse compressed features. (c) depicts the feature importance scores, represented by the learnable feature weight $\pi$ in the Gumbel-Softmax function. Mutual information exceeding a specific threshold is highlighted with a red box in (a) and (b).

From the perspective of selective retention of informative features, only a small subset of input features exhibits high mutual information with the latent space. Specifically, in the perceptual compression plot, we observe that around 15\%-20\% of the input features have mutual information values above 0.15, indicating that these features are essential for retaining critical information. Similarly, in the sparse compression plot, about 10\%-15\% of the input features show mutual information values above 0.20. This indicates effective compression by both ATA and DGFS, as redundant information is removed, retaining only the most informative features. Certain input features, such as those indexed around 8, 22, and 45, consistently show high mutual information across multiple latent features in both perceptual and sparse compressed plots. These features are crucial as they maintain strong correlations with various aspects of the latent space, ensuring that the model retains essential information.

The \hyperref[fig:f2]{Fig.~\ref{fig:f2}(c)} reveals that features with higher mutual information values also tend to have higher importance scores. These importance scores are derived from the feature weights learned through the Gumbel-Softmax function, which assigns higher weights to features that contribute more significantly to the model's predictive accuracy. For instance, the feature indexed at 11, which shows high mutual information, also has a high importance score of approximately 0.03. Similarly, the feature indexed at 66 has an importance score of about 0.075. This overlap between mutual information and feature importance score indicates that the features identified as important by mutual information metrics are indeed critical for the model’s predictive accuracy. This reinforces the robustness of our feature selection process, ensuring that the most informative features drive the model’s predictions.

\subsection{Interpretability and Clinical Insights}
To enhance the interpretability of the model, we have summarized and categorized the clinical significance of the selected features for the mortality prediction task and listed them in \hyperref[tab:feature_significance]{Table~\ref{tab:feature_significance}}.

\begin{table}
    \centering
    \caption{Summary of selected features and their clinical significance}
    \resizebox{\textwidth}{!}{
        \begin{tabular}{cc}
        \toprule
         \textbf{Feature} & \textbf{Clinical Significance} \\
         \midrule
         Respiratory rate & Indicates the patient's breathing status\\
         GCS Verbal Response (5 Oriented) & Reflects the patient's cognitive function and orientation. \\
         GCS Verbal Response (1.0 ET/Trach) & Indicates the presence of intubation or tracheostomy, preventing verbal communication. \\
         GCS Eye Opening (4 Spontaneously) & Represents a high level of consciousness, indicating a responsive patient. \\
         GCS Motor Response (Localizes Pain) & Reflects the patient's ability to localize pain, important for assessing neurological function. \\
         GCS Total & Provides an overall assessment of consciousness and neurological function. \\
         mask$\rightarrow$Glucose & Critical for assessing metabolic status\\
         \bottomrule
    \end{tabular}
    }
    \label{tab:feature_significance}
\end{table}


These features include vital signs, neurological assessments, and metabolic indicators, all of which are crucial for assessing a patient’s overall health condition and are strongly linked to mortality outcomes. The clinical relevance of these features highlights their importance in guiding the model's predictions, ensuring that the decision-making process is not only data-driven but also aligned with real-world medical understanding, thus making it more transparent and interpretable for medical professionals.

\begin{figure}
    \centering
    \includegraphics[width=1\linewidth]{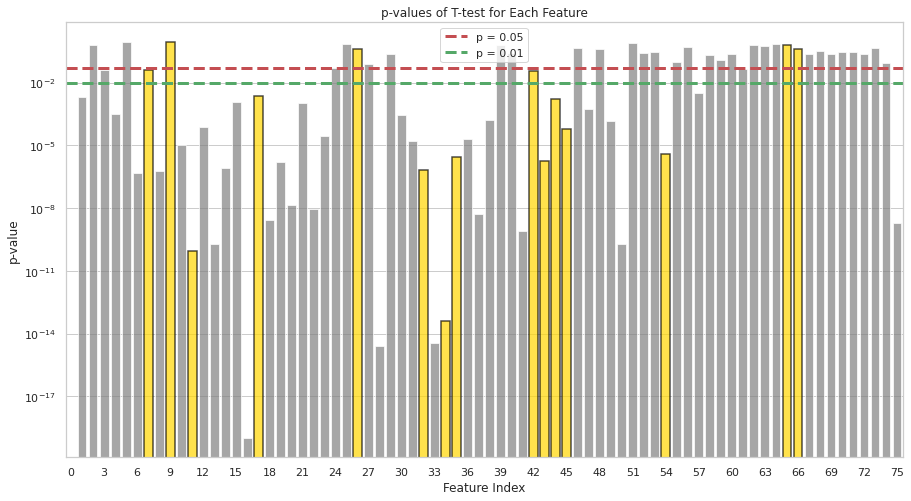}
    \caption{The p-values of T-test for each input feature. Features selected by the DGFS module are highlighted in yellow. The significance thresholds are marked by dashed lines: green for $p < 0.01$ and red for $p < 0.05$.}
    \label{fig:f3}
\end{figure}

To further validate the importance of the selected features, we conducted hypothesis testing using T-test on all input features. The T-test assesses whether the differences between the means of selected features and the rest of the dataset are statistically significant. A lower p-value indicates greater statistical significance. As shown in \hyperref[fig:f3]{Fig.~\ref{fig:f3}}, the majority of the features selected by the DGFS module have p-values below 0.01, indicating their high statistical significance. For instance, features indexed at 11 and 34 have p-values less than $10^{-9}$, highlighting their substantial importance in the model's predictions. Conversely, many non-selected features exhibit higher p-values, often exceeding the 0.05 threshold, suggesting they are less critical for the predictive model. Overall, the hypothesis testing results robustly validate the DGFS module's effectiveness in selecting features that are both clinically and statistically significant, enhancing the model's interpretability and reliability.

\subsection{Predictive Performance and Comparison}
\begin{table}
    \centering
    \caption{Results and comparison of mortality prediction on the MIMIC-III dataset.}
    \begin{tabular}{cccc}
        \toprule
         \multicolumn{4}{c}{Mortality prediction on MIMIC-III dataset}\\
         Methods &  AUROC &  AUPRC & min(Se,P+) 
\\      \midrule
         GRU$_{\alpha}$&  .8628(.011) &  .4989(.022) & .5026(.028) 
\\
         RETAIN &  .8313(.014) &  .4790(.020) & .4721(.022) 
\\
         T-LSTM &  .8617(.014) &  .4964(.022) & .4977(.029) 
\\
         SAnD* &  .8382(.007) &  .4545(.018) & .4885(.017) 
\\
         ConCare &  .8702(.008) &  .5317(.027) & .5082(.021) 
\\
         DeepSelective &  \textbf{.9054(.013)} &  \textbf{.5627(.019)} & \textbf{.5218(.020)} \\
         \bottomrule
          \multicolumn{4}{c}{Decompensation Prediction of MIMIC }\\
          &  AUROC & AUPRC &  min(Se,P+) 
\\        \midrule
          GRU$_{\alpha}$ & .8983(.003) &  .2784(.003) &  .3260(.004)
\\
          RETAIN & .8764(.002) &  .2597(.004) &  .2900(.005) 
\\
          T-LSTM & .8944(.002) &  .2611(.003) &  .3186(.004)
\\
          SAnD* & .8825(.003) &  .2524(.003) &  .2899(.004)
\\
          AdaCare & .9004(.003) &  .3037(.004) &  .3429(.004)
\\
          DeepSelective & \textbf{.9143(.003)} &  \textbf{.3203(.003)} &  \textbf{.3573(.004)} \\
          \bottomrule
         
    \end{tabular}
    \label{tab:t1}
\end{table}


This section presents a detailed comparison of our proposed DeepSelective framework against several baseline methods across two datasets: mortality and decompensation prediction. The quantitative results are summarized in \hyperref[tab:t1]{Table~\ref{tab:t1}}.

DeepSelective achieves an AUROC of 0.9054 and an AUPRC of 0.5627, outperforming other models such as GRU$_{\alpha}$, RETAIN, and T-LSTM. Notably, DeepSelective’s min(Se, P+) is 0.5218, which is also superior to other models. For the decompensation prediction task, DeepSelective achieves an AUROC of 0.9143, an AUPRC of 0.3203, and a min(Se, P+) of 0.3573, which is also the best model. The experimental results demonstrate that our DeepSelective framework generally outperforms baseline models in terms of AUPRC, AUROC, and min(Se, P+), solidifying its position as a leading tool for predictive modeling in EHR-Based Prognosis Prediction.

\subsection{Latent Space Exploration}
In this section, we delve into the analysis of the latent space representations generated by our framework. Understanding the latent space is crucial as it provides insight into how well the model captures the underlying structure of the data. We employ principal component analysis (PCA)\citep{48} to visualize the latent space of input features, perceptually compressed features, and sparse compressed features.

\begin{figure}
    \centering
    \includegraphics[width=1\linewidth]{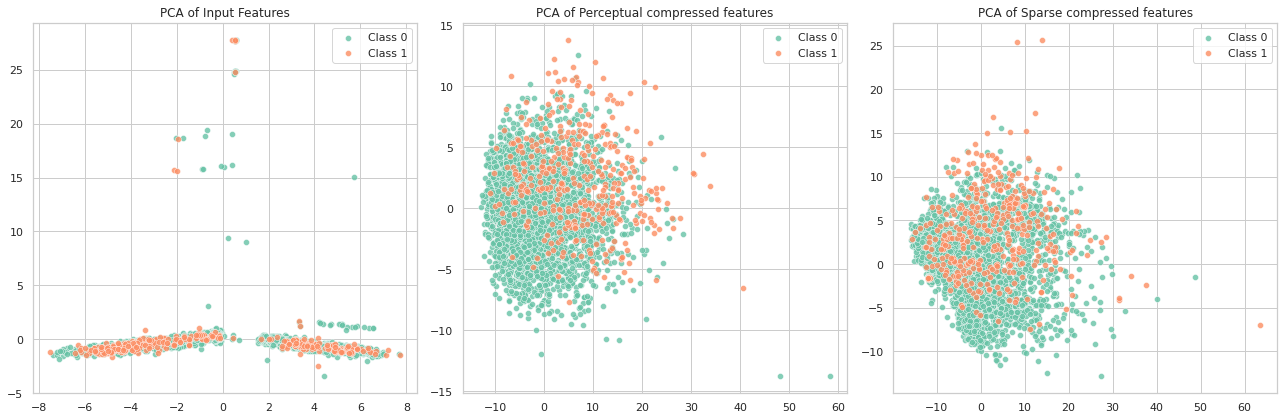}
    \caption{PCA visualizations of different feature sets. (a) shows the PCA of the raw input features, where Class 0 (green) and Class 1 (blue) exhibit significant overlap, indicating poor class separation. (b) displays the PCA of perceptual compressed features derived from the Attentive Transformer Autoencoder (ATA) and (c) illustrates the PCA of sparse compressed features obtained from the Dynamic Gating Feature Selection (DGFS) module.}
    \label{fig:pca}
\end{figure}

In the PCA of the raw input features in \hyperref[fig:pca]{Fig.~\ref{fig:pca} (a)}, we observe significant overlap between the data points of class 0 (green) and class 1 (blue), indicating poor class separability. This suggests that the input space contains a considerable amount of redundant or less informative features, which hinders the model’s ability to distinguish between classes effectively. In contrast, the PCA of the perceptual compressed features \hyperref[fig:pca]{Fig.~\ref{fig:pca} (b)} and the sparse compressed features \hyperref[fig:pca]{Fig.~\ref{fig:pca}(c)} shows much better class separation. The perceptual compressed features, derived from the Attentive Transformer Autoencoder (ATA), form more compact and distinct clusters for each class, demonstrating that the ATA module effectively captures and retains the most relevant information. Similarly, the sparse compressed features obtained from the dynamic gate feature selection module (DGFS) exhibit a clear class separation with minimal overlap, highlighting the module’s ability to identify and preserve the most critical features. In particular, the ATA module appears to provide better separation than the DGFS, which could be attributed to the powerful representational learning capabilities of deep learning. This observation supports one of the main hypotheses of our study: that the robust feature extraction capabilities of deep learning significantly improve the quality of the representation of the latent space. In general, these visualizations confirm that both compression methods improve the model's ability to differentiate between classes, leading to improved predictive performance and interpretability.

\subsection{Ablation Study}

\begin{table}
    \centering
    \caption{Component configuration for the DeepSelective framework and its ablated variants.}
    \begin{tabular}{ccccc}
    \toprule
                & DGFS &  ATA &  SparsityController & RML \\
    \midrule
         DeepSelective & $\checkmark$ & $\checkmark$  & $\checkmark$ & $\checkmark$\\
         $\text{DeepSelective}_{na}$ & $\checkmark$ & & $\checkmark$ \\
         $\text{DeepSelective}_{nd}$ & & $\checkmark$ & \\
         $\text{DeepSelective}_{nc}$ & $\checkmark$ & $\checkmark$ & & $\checkmark$ \\
         \bottomrule
    \end{tabular}
    \label{tab:t3}
\end{table}

The  conducted study on the MIMIC-III dataset highlights the significance of each component within the DeepSelective framework. We evaluated the full DeepSelective model against three variants: $\text{DeepSelective}_{na}$, $\text{DeepSelective}_{nd}$, and $\text{DeepSelective}_{nc}$. The detailed configuration for each model can be seen in \hyperref[tab:t3]{Table~\ref{tab:t3}}. The performance metrics considered were AUROC, AUPRC, and min(Se, P+), with results presented for both mortality and decompensation prediction tasks.

\begin{figure}
    \centering
    \includegraphics[width=1\linewidth]{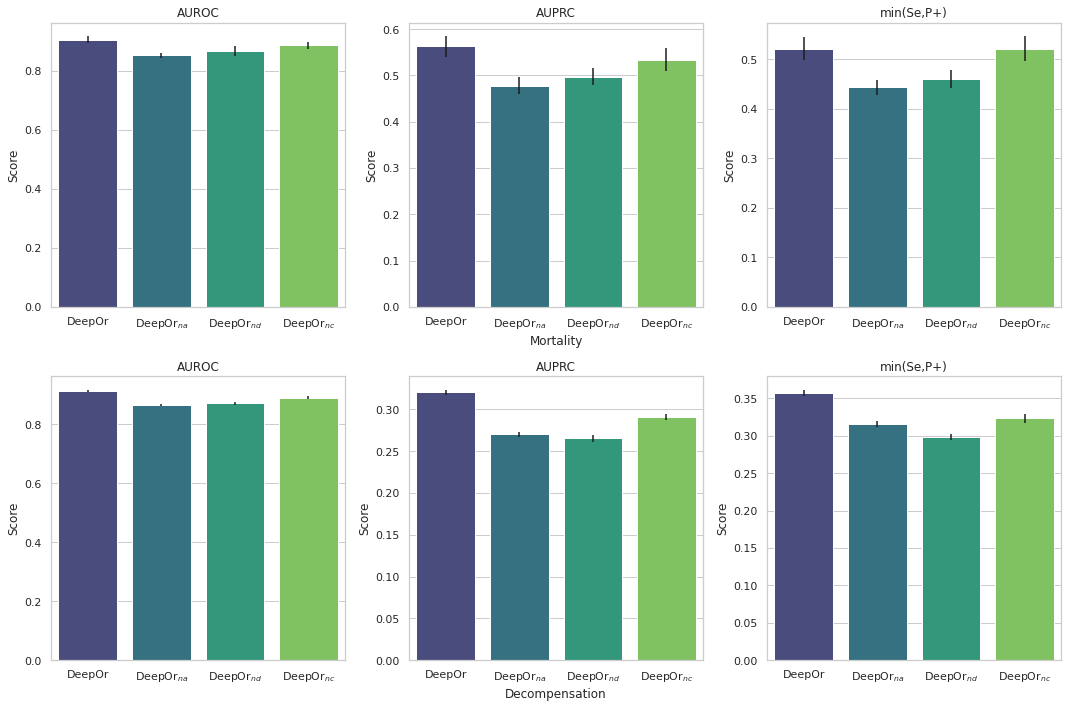}
    \caption{The  results of the DeepSelective framework compared to its variants without certain components.}
    \label{fig:f4}
\end{figure}

As visualized in the \hyperref[fig:f4]{Fig.~\ref{fig:f4}}, the full DeepSelective model consistently outperforms its ablated variants across all metrics. For mortality prediction, DeepSelective achieves an AUROC of 0.9054, an AUPRC of 0.5627, and a min(Se, P+) of 0.5218. In contrast, the removal of the ATA module results in the most substantial decline in performance, with DeepSelective\_na showing an AUROC of 0.8524 and an AUPRC of 0.4781. This significant drop underscores the crucial role of the ATA module in capturing and retaining essential information through its powerful representational learning capabilities. The DGFS module and SparsityController also contribute significantly to the model’s performance, as evidenced by the decreased scores in their respective ablated versions. Similar trends are observed in the decompensation prediction task, where DeepSelective attains an AUROC of 0.9143, an AUPRC of 0.3203, and a min(Se, P+) of 0.3573. The performance declines in the ablated models further emphasize the importance of each component.

Overall, the  demonstrates that the integrated design of the DeepSelective framework, which combines the strengths of the DGFS, ATA and SparsityController modules, is crucial for achieving superior predictive performance and interpretability. The substantial performance drops in the ablated models validate our hypothesis that the synergistic combination of these modules enables DeepSelective to effectively handle high-dimensional EHRs data, capture relevant features, and provide meaningful insights into the prediction process. 

\section{Conclusion}
In this work, we presented DeepSelective, a novel end-to-end deep learning framework tailored for clinical prognosis prediction using high-dimensional EHR data. The framework introduces a dual compression design: a Dynamic Gating Feature Selection (DGFS) module for sparse, interpretable feature selection, and an Attentive Transformer Autoencoder (ATA) for perceptual compression via deep representation learning. Combined with a Representation Matching Layer (RML) and a SparsityController, DeepSelective effectively balances model performance and interpretability, two often conflicting goals in clinical AI applications. 

Our extensive experiments on two benchmark clinical prediction tasks—mortality prediction and real-time decompensation detection—demonstrate that DeepSelective consistently outperforms state-of-the-art baselines across multiple metrics (AUROC, AUPRC, min(Se,P+)). Furthermore, the ablation studies confirm the necessity of each component within the framework, showing clear performance degradation when any module is removed. We also validated the clinical meaningfulness of selected features through mutual information analysis, T-tests, and PCA visualizations, highlighting the model’s transparency and relevance to real-world clinical decision-making. 

However, our framework has limitations. First, the temperature tuning in the Gumbel-Softmax-based SparsityController is currently guided by heuristic initialization and PID-based adjustments, which might not generalize optimally across all data distributions. Second, although our evaluation on MIMIC-III datasets is thorough, external validation on broader EHR systems—across hospitals, disease domains, and demographic groups—remains to be conducted. In addition, model performance could potentially degrade in the presence of severe missingness or noisy features not addressed by preprocessing alone.

We believe DeepSelective contributes a principled and practical approach to interpretable prognosis modeling. It offers modular components (e.g., plug-in feature selectors, matching layers) that other researchers can adapt for various structured data prediction problems beyond healthcare, such as finance or manufacturing. Moreover, clinicians and data scientists can benefit from its transparent prediction process, where key features and representations are explicitly surfaced. In future work, we plan to develop self-adaptive sparsity control mechanisms using reinforcement learning or meta-learning to replace manual tuning, incorporate multi-modal clinical data such as medical imaging and unstructured notes to improve representation learning, and explore the use of counterfactual explanations to further enhance model interpretability.

\section*{Acknowledgments}
This work was supported by the National Natural Science Foundation of China (No. 62072212), Development Project of Jilin Province of China (No. 20220508125RC), Guizhou Provincial Science and Technology Projects (ZK2023-297), the Science and Technology Foundation of Health Commission of Guizhou Province (gzwkj2023-565), and the Fundamental Research Funds for the Central Universities (JLU).

 \bibliographystyle{elsarticle-num} 
 \bibliography{main}

\begin{thebibliography}{10}
\expandafter\ifx\csname url\endcsname\relax
  \def\url#1{\texttt{#1}}\fi
\expandafter\ifx\csname urlprefix\endcsname\relax\def\urlprefix{URL }\fi
\expandafter\ifx\csname href\endcsname\relax
  \def\href#1#2{#2} \def\path#1{#1}\fi

\bibitem{vaswani2017attention}
A.~Vaswani, N.~Shazeer, N.~Parmar, J.~Uszkoreit, L.~Jones, A.~N. Gomez, {\L}.~Kaiser, I.~Polosukhin, Attention is all you need, Advances in neural information processing systems 30 (2017).

\bibitem{nasarudin2024review}
N.~A. Nasarudin, F.~Al~Jasmi, R.~O. Sinnott, N.~Zaki, H.~Al~Ashwal, E.~A. Mohamed, M.~S. Mohamad, A review of deep learning models and online healthcare databases for electronic health records and their use for health prediction, Artificial Intelligence Review 57~(9) (2024) 249.

\bibitem{willis1999proportional}
M.~J. Willis, Proportional-integral-derivative control, Dept. of Chemical and Process Engineering University of Newcastle 6 (1999).

\bibitem{li2022interpretable}
X.~Li, H.~Xiong, X.~Li, X.~Wu, X.~Zhang, J.~Liu, J.~Bian, D.~Dou, Interpretable deep learning: Interpretation, interpretability, trustworthiness, and beyond, Knowledge and Information Systems 64~(12) (2022) 3197--3234.

\bibitem{mulyani2024analysis}
S.~Mulyani, Analysis of the impact of electronic health record use on the effectiveness of diagnostic and treatment processes, The Journal of Academic Science 1~(8) (2024) 931--941.

\bibitem{jang2016categorical}
E.~Jang, S.~Gu, B.~Poole, Categorical reparameterization with gumbel-softmax, in: International Conference on Learning Representations (ICLR), 2017, arXiv preprint arXiv:1611.01144.

\bibitem{2}
R.~Fang, S.~Pouyanfar, Y.~Yang, S.-C. Chen, S.~Iyengar, Computational health informatics in the big data age: a survey, ACM Computing Surveys (CSUR) 49~(1) (2016) 1--36.

\bibitem{sittig2025recommendations}
D.~F. Sittig, H.~Singh, Recommendations to ensure safety of ai in real-world clinical care, JAMA 333~(6) (2025) 457--458.

\bibitem{4}
R.~Garriga, J.~Mas, S.~Abraha, J.~Nolan, O.~Harrison, G.~Tadros, A.~Matic, Machine learning model to predict mental health crises from electronic health records, Nature medicine 28~(6) (2022) 1240--1248.

\bibitem{karaferis2025design}
D.~Karaferis, D.~Balaska, Y.~Pollalis, Design and development of data-driven ai to reduce the discrepancies in healthcare ehr utilization, J Clin Med Re: AJCMR-184 (2025).

\bibitem{zhang2024multi}
Y.~Zhang, W.~Huo, J.~Tang, Multi-label feature selection via latent representation learning and dynamic graph constraints, Pattern Recognition 151 (2024) 110411.

\bibitem{li2024isp}
X.~Li, Y.~Yang, H.~Zheng, Z.~Xu, Isp-irlnet: Joint optimization of interpretable sampler and implicit regularization learning network for accerlerated mri, Pattern Recognition 151 (2024) 110412.

\bibitem{zhang2025brain}
L.~Zhang, J.~Wu, L.~Wang, L.~Wang, D.~C. Steffens, S.~Qiu, G.~G. Potter, M.~Liu, Brain anatomy prior modeling to forecast clinical progression of cognitive impairment with structural mri, Pattern Recognition 165 (2025) 111603.

\bibitem{6}
C.~Castaneda, K.~Nalley, C.~Mannion, P.~Bhattacharyya, P.~Blake, A.~Pecora, A.~Goy, K.~S. Suh, Clinical decision support systems for improving diagnostic accuracy and achieving precision medicine, Journal of clinical bioinformatics 5 (2015) 1--16.

\bibitem{7}
B.~Shickel, P.~J. Tighe, A.~Bihorac, P.~Rashidi, Deep ehr: a survey of recent advances in deep learning techniques for electronic health record (ehr) analysis, IEEE journal of biomedical and health informatics 22~(5) (2017) 1589--1604.

\bibitem{8}
F.~Liu, C.~Weng, H.~Yu, Advancing clinical research through natural language processing on electronic health records: traditional machine learning meets deep learning, Clinical Research Informatics (2019) 357--378.

\bibitem{9}
Y.~Li, S.~Rao, J.~R.~A. Solares, A.~Hassaine, R.~Ramakrishnan, D.~Canoy, Y.~Zhu, K.~Rahimi, G.~Salimi-Khorshidi, Behrt: transformer for electronic health records, Scientific reports 10~(1) (2020) 7155.

\bibitem{10}
Y.~Wang, R.~Zhang, Q.~Yang, Q.~Zhou, S.~Zhang, Y.~Fan, L.~Huang, K.~Li, F.~Zhou, Faircare: Adversarial training of a heterogeneous graph neural network with attention mechanism to learn fair representations of electronic health records, Information Processing \& Management 61~(3) (2024) 103682.

\bibitem{11}
R.~Miotto, F.~Wang, S.~Wang, X.~Jiang, J.~T. Dudley, Deep learning for healthcare: review, opportunities and challenges, Briefings in bioinformatics 19~(6) (2018) 1236--1246.

\bibitem{12}
J.~Gronsbell, J.~Minnier, S.~Yu, K.~Liao, T.~Cai, Automated feature selection of predictors in electronic medical records data, Biometrics 75~(1) (2019) 268--277.

\bibitem{13}
A.~Rios, R.~Kavuluru, Supervised extraction of diagnosis codes from emrs: role of feature selection, data selection, and probabilistic thresholding, in: 2013 IEEE International Conference on Healthcare Informatics, IEEE, 2013, pp. 66--73.

\bibitem{14}
R.~J. Carroll, A.~E. Eyler, J.~C. Denny, Na{\"\i}ve electronic health record phenotype identification for rheumatoid arthritis, in: AMIA annual symposium proceedings, Vol. 2011, American Medical Informatics Association, 2011, p. 189.

\bibitem{15}
H.~Li, X.~Li, X.~Jia, M.~Ramanathan, A.~Zhang, Bone disease prediction and phenotype discovery using feature representation over electronic health records, in: Proceedings of the 6th ACM Conference on Bioinformatics, Computational Biology and Health Informatics, 2015, pp. 212--221.

\bibitem{16}
C.~Xiao, E.~Choi, J.~Sun, Opportunities and challenges in developing deep learning models using electronic health records data: a systematic review, Journal of the American Medical Informatics Association 25~(10) (2018) 1419--1428.

\bibitem{17}
S.~Chakraborty, R.~Tomsett, R.~Raghavendra, D.~Harborne, M.~Alzantot, F.~Cerutti, M.~Srivastava, A.~Preece, S.~Julier, R.~M. Rao, et~al., Interpretability of deep learning models: A survey of results, in: 2017 IEEE smartworld, ubiquitous intelligence \& computing, advanced \& trusted computed, scalable computing \& communications, cloud \& big data computing, Internet of people and smart city innovation (smartworld/SCALCOM/UIC/ATC/CBDcom/IOP/SCI), IEEE, 2017, pp. 1--6.

\bibitem{18}
J.~Zeng, M.~F. Gensheimer, D.~L. Rubin, S.~Athey, R.~D. Shachter, Uncovering interpretable potential confounders in electronic medical records, Nature communications 13~(1) (2022) 1014.

\bibitem{19}
X.~Liu, Y.~Zhang, C.~Fu, R.~Zhang, F.~Zhou, Enrank: an ensemble method to detect pulmonary hypertension biomarkers based on feature selection and machine learning models, Frontiers in Genetics 12 (2021) 636429.

\bibitem{20}
M.~Badawy, N.~Ramadan, H.~A. Hefny, Healthcare predictive analytics using machine learning and deep learning techniques: a survey, Journal of Electrical Systems and Information Technology 10~(1) (2023) 40.

\bibitem{21}
J.~Pathak, A.~N. Kho, J.~C. Denny, Electronic health records-driven phenotyping: challenges, recent advances, and perspectives, Journal of the American Medical Informatics Association 20~(e2) (2013) e206--e211.

\bibitem{22}
S.~Uddin, A.~Khan, M.~E. Hossain, M.~A. Moni, Comparing different supervised machine learning algorithms for disease prediction, BMC medical informatics and decision making 19~(1) (2019) 1--16.

\bibitem{23}
A.~Akbulut, E.~Ertugrul, V.~Topcu, Fetal health status prediction based on maternal clinical history using machine learning techniques, Computer methods and programs in biomedicine 163 (2018) 87--100.

\bibitem{24}
K.~Shameer, K.~W. Johnson, A.~Yahi, R.~Miotto, L.~Li, D.~Ricks, J.~Jebakaran, P.~Kovatch, P.~P. Sengupta, S.~Gelijns, et~al., Predictive modeling of hospital readmission rates using electronic medical record-wide machine learning: a case-study using mount sinai heart failure cohort, in: Pacific symposium on biocomputing 2017, World Scientific, 2017, pp. 276--287.

\bibitem{25}
Y.~K. Wan, C.~Hendra, P.~N. Pratanwanich, J.~G{\"o}ke, Beyond sequencing: machine learning algorithms extract biology hidden in nanopore signal data, Trends in Genetics 38~(3) (2022) 246--257.

\bibitem{26}
L.~Ma, J.~Gao, Y.~Wang, C.~Zhang, J.~Wang, W.~Ruan, W.~Tang, X.~Gao, X.~Ma, Adacare: Explainable clinical health status representation learning via scale-adaptive feature extraction and recalibration, in: Proceedings of the AAAI Conference on Artificial Intelligence, Vol.~34, 2020, pp. 825--832.

\bibitem{27}
C.~Zhang, X.~Gao, L.~Ma, Y.~Wang, J.~Wang, W.~Tang, Grasp: generic framework for health status representation learning based on incorporating knowledge from similar patients, in: Proceedings of the AAAI conference on artificial intelligence, Vol.~35, 2021, pp. 715--723.

\bibitem{28}
O.~Loyola-Gonzalez, Black-box vs. white-box: Understanding their advantages and weaknesses from a practical point of view, IEEE access 7 (2019) 154096--154113.

\bibitem{29}
S.~V. Kovalchuk, G.~D. Kopanitsa, I.~V. Derevitskii, G.~A. Matveev, D.~A. Savitskaya, Three-stage intelligent support of clinical decision making for higher trust, validity, and explainability, Journal of Biomedical Informatics 127 (2022) 104013.

\bibitem{30}
C.~J. Kelly, A.~Karthikesalingam, M.~Suleyman, G.~Corrado, D.~King, Key challenges for delivering clinical impact with artificial intelligence, BMC medicine 17 (2019) 1--9.

\bibitem{31}
P.~Linardatos, V.~Papastefanopoulos, S.~Kotsiantis, Explainable ai: A review of machine learning interpretability methods, Entropy 23~(1) (2020) 18.

\bibitem{32}
H.-T. Cheng, L.~Koc, J.~Harmsen, T.~Shaked, T.~Chandra, H.~Aradhye, G.~Anderson, G.~Corrado, W.~Chai, M.~Ispir, et~al., Wide \& deep learning for recommender systems, in: Proceedings of the 1st workshop on deep learning for recommender systems, 2016, pp. 7--10.

\bibitem{33}
Y.~Huang, W.~Jin, Z.~Yu, B.~Li, Supervised feature selection through deep neural networks with pairwise connected structure, Knowledge-Based Systems 204 (2020) 106202.

\bibitem{34}
V.~Borisov, J.~Haug, G.~Kasneci, Cancelout: A layer for feature selection in deep neural networks, in: Artificial Neural Networks and Machine Learning--ICANN 2019: Deep Learning: 28th International Conference on Artificial Neural Networks, Munich, Germany, September 17--19, 2019, Proceedings, Part II 28, Springer, 2019, pp. 72--83.

\bibitem{35}
E.~Choi, M.~T. Bahadori, J.~Sun, J.~Kulas, A.~Schuetz, W.~Stewart, Retain: An interpretable predictive model for healthcare using reverse time attention mechanism, Advances in neural information processing systems 29 (2016).

\bibitem{36}
I.~M. Baytas, C.~Xiao, X.~Zhang, F.~Wang, A.~K. Jain, J.~Zhou, Patient subtyping via time-aware lstm networks, in: Proceedings of the 23rd ACM SIGKDD international conference on knowledge discovery and data mining, 2017, pp. 65--74.

\bibitem{37}
P.~Gupta, P.~Malhotra, L.~Vig, G.~Shroff, Using features from pre-trained timenet for clinical predictions., in: KDH@ IJCAI, 2018, pp. 38--44.

\bibitem{38}
H.~Song, D.~Rajan, J.~Thiagarajan, A.~Spanias, Attend and diagnose: Clinical time series analysis using attention models, in: Proceedings of the AAAI conference on artificial intelligence, Vol.~32, 2018.

\bibitem{39}
L.~Ma, C.~Zhang, Y.~Wang, W.~Ruan, J.~Wang, W.~Tang, X.~Ma, X.~Gao, J.~Gao, Concare: Personalized clinical feature embedding via capturing the healthcare context, in: Proceedings of the AAAI Conference on Artificial Intelligence, Vol.~34, 2020, pp. 833--840.

\bibitem{40}
M.~Singer, C.~S. Deutschman, C.~W. Seymour, M.~Shankar-Hari, D.~Annane, M.~Bauer, R.~Bellomo, G.~R. Bernard, J.-D. Chiche, C.~M. Coopersmith, et~al., The third international consensus definitions for sepsis and septic shock (sepsis-3), Jama 315~(8) (2016) 801--810.

\bibitem{41}
M.~A. Reyna, C.~S. Josef, R.~Jeter, S.~P. Shashikumar, M.~B. Westover, S.~Nemati, G.~D. Clifford, A.~Sharma, Early prediction of sepsis from clinical data: the physionet/computing in cardiology challenge 2019, Critical care medicine 48~(2) (2020) 210--217.

\bibitem{42}
G.~B. Moody, R.~G. Mark, A database to support development and evaluation of intelligent intensive care monitoring, in: Computers in Cardiology 1996, IEEE, 1996, pp. 657--660.

\bibitem{43}
A.~E. Johnson, T.~J. Pollard, L.~Shen, L.-w.~H. Lehman, M.~Feng, M.~Ghassemi, B.~Moody, P.~Szolovits, L.~Anthony~Celi, R.~G. Mark, Mimic-iii, a freely accessible critical care database, Scientific data 3~(1) (2016) 1--9.

\bibitem{44}
H.~Harutyunyan, H.~Khachatrian, D.~C. Kale, G.~Ver~Steeg, A.~Galstyan, Multitask learning and benchmarking with clinical time series data, Scientific data 6~(1) (2019) 96.

\bibitem{45}
L.~Ma, J.~Gao, Y.~Wang, C.~Zhang, J.~Wang, W.~Ruan, W.~Tang, X.~Gao, X.~Ma, Adacare: Explainable clinical health status representation learning via scale-adaptive feature extraction and recalibration, in: Proceedings of the AAAI Conference on Artificial Intelligence, Vol.~34, 2020, pp. 825--832.

\bibitem{46}
A.~Kraskov, H.~St{\"o}gbauer, P.~Grassberger, Estimating mutual information, Physical Review E—Statistical, Nonlinear, and Soft Matter Physics 69~(6) (2004) 066138.

\bibitem{47}
Y.~Ye, R.~Zhang, W.~Zheng, S.~Liu, F.~Zhou, Rifs: a randomly restarted incremental feature selection algorithm, Scientific Reports 7~(1) (2017) 13013.

\bibitem{48}
S.~Wold, K.~Esbensen, P.~Geladi, Principal component analysis, Chemometrics and intelligent laboratory systems 2~(1-3) (1987) 37--52.

\end{thebibliography}





\end{document}